\def\year{2021}\relax
\documentclass[letterpaper]{article} 
\usepackage{aaai21}  
\usepackage{times}  
\usepackage{helvet} 
\usepackage{courier}  
\usepackage[hyphens]{url}  
\usepackage{graphicx} 
\usepackage{natbib}  
\usepackage{caption} 
\usepackage[switch]{lineno}
\usepackage{times}
\usepackage{multirow}
\usepackage{latexsym}

\usepackage{amsmath}
\usepackage{bm}
\usepackage{verbatim}
\usepackage{graphicx}
\usepackage{subfigure}
\usepackage{diagbox}
\usepackage{amssymb}

\title{Future-Guided Incremental Transformer for Simultaneous Translation}
\author{
    Shaolei Zhang \textsuperscript{\rm 1,2},
    Yang Feng \textsuperscript{\rm 1,2}\thanks{Corresponding author: Yang Feng.},
    Liangyou Li\textsuperscript{\rm 3}\\
}
\affiliations{
    \textsuperscript{\rm 1}{Key Laboratory of Intelligent Information Processing} \\ Institute of Computing Technology, Chinese Academy of Sciences (ICT/CAS) \\
    { \textsuperscript{\rm 2} {University of Chinese Academy of Sciences, Beijing, China}} \\
    \textsuperscript{\rm 3} {Huawei Noah's Ark Lab}\\
    \{{zhangshaolei20z, fengyang}\}@ ict.ac.cn\\
     liliangyou@huawei.com

}

\begin{document}

\maketitle

\begin{abstract}
Simultaneous translation (ST) starts translations synchronously while reading source sentences, and is used in many online scenarios. The previous wait-k policy is concise and achieved good results in ST. However, wait-k policy faces two weaknesses: low training speed caused by the recalculation of hidden states and lack of future source information to guide training. For the low training speed, we propose an incremental Transformer with an average embedding layer (AEL) to accelerate the speed of calculation of the hidden states during training. For future-guided training, we propose a conventional Transformer as the teacher of the incremental Transformer, and try to invisibly embed some future information in the model through knowledge distillation. We conducted experiments on Chinese-English and German-English simultaneous translation tasks and compared with the wait-k policy to evaluate the proposed method. Our method can effectively increase the training speed by about 28 times on average at different $k$ and implicitly embed some predictive abilities in the model, achieving better translation quality than wait-k baseline.
\end{abstract}

\section{Introduction}
Simultaneous translation(ST) \cite{Cho2016,gu-etal-2017-learning,ma-etal-2019-stacl,Arivazhagan2019}, a variant of machine translation, aims to output the translations while reading source sentences, which is more suitable for input-output synchronization tasks (such as online translation, live subtitle and simultaneous interpretation). 

Recently, wait-k policy \cite{ma-etal-2019-stacl} is a widely used read / write policy, which first waits for $k$ source tokens, and then translates concurrently with the rest of source sentence. Wait-k policy was trained by a ``prefix-to-prefix'' architecture, and need to recalculate the hidden states of all previous source tokens when a new source token is received. The wait-k policy achieved excellent results in ST and successfully integrated some implicit anticipation through ``prefix-to-prefix'' training. 

However, there are still two shortcomings in the adoption of source information. According to whether the token is read, all source tokens can be divided into two categories: consumed and future. First, for the consumed source, wait-k needs to re-calculate the hidden states of all previous source tokens at each decoding step, making the computational cost increase quadratically \cite{dalvi-etal-2018-incremental,chen2020general}. The growth factor of the computational cost in training is proportional to the length of the target sentence. 
Second, for the future source, since wait-k policy is trained with ``prefix-to-prefix'' architecture, some source tokens will lag behind due to the different word order, which is not considered in training. Although ``prefix-to-prefix'' architecture makes wait-k policy have some implicit anticipation, \citet{ma-etal-2019-stacl} pointed that the acquisition of implicit anticipation is data-driven, since the training data contains many prefix-pairs in the similar form. We consider that the data-driven approach is inefficient and uncontrollable. During training, wait-k policy lacks the guidance from future source information, to gain a stronger predictive ability.

To address the above two problems, we propose a \textbf{Future-Guided Incremental Transformer} with average embedding layer (AEL) and knowledge distillation \cite{kn}. The proposed method greatly accelerate the training speed, meanwhile plenty exploit the future information to guide training and enable the model to obtain a stronger predictive ability. 

To avoid the high complexity caused by recalculation of the consumed source hidden states, inspired by \citet{zhang-etal-2018-accelerating}, we propose the incremental Transformer, including a unidirectional encoder and a decoder with an average embedding layer. The average embedding layer is added into decoder to summarize the consumed source information, by calculating the average embedding of all consumed source tokens. Therefore, each token can attend to all consumed tokens through the unidirectional encoder and AEL, avoiding the recalculation at the same time.


To utilize future source information to enhance the predictive ability, we encourage the model to embed some future information through knowledge distillation \cite{kn,Ravanelli2018,Novitasari2019}. Unlike some previous methods of adding `predict operation' to ST, out method do not explicitly predict the next word or verb, but implicitly embed the future information in the model. While training incremental Transformer (student), we simultaneously trained a conventional Transformer for full-sentence NMT as the teacher of incremental Transformer. Thus, the incremental Transformer can learn some future information from the conventional Transformer. While testing, we only use incremental Transformer for ST, so that it does not introduce any waiting time or any calculations.

Experiment results on the Chinese-English, German-English simultaneous translation tasks show our method outperforms the baseline.

In summary, our contributions are two-fold:
\begin{itemize}\setlength{\itemsep}{5pt}
\item Our method does not need to recalculate the hidden states of encoder, and also allows each source token to attend to the complete consumed source. In training, our method can greatly accelerate the training speed about 28 times.
\item Our method provides a way to embed future information in the incremental model, and effectively enhances the predictive ability of the incremental model without adding any waiting time or parameters during the inference time.
\end{itemize}

\section{Background}
We propose our method based on full-sentence NMT and wait-k policy \cite {ma-etal-2019-stacl}, so we first briefly introduce them.

\subsection{Full-Sentence NMT}
Transformer \cite{NIPS2017_7181} is currently the most widely used model for full-sentence NMT. Transformer consists of two parts, encoder and decoder, each of which contains $N$ repeated independent structures. 
The input sentence is $\mathbf{x}=\left ( x_{1} , \cdots ,x_{n}\right )$, where $x_{i}\in \mathbb{R}^{d_{model}}$ and $d_{model}$ represents the representation dimension. The encoder maps $\mathbf{x}$ to a sequence of hidden states $\mathbf{z}=\left ( z_{1} , \cdots ,z_{n}\right )$. Given $\mathbf{z}$ and the previous target tokens, the decoder predicts the next output token $y_{t}$, and finally the entire output sequence is $\mathbf{y}=\left ( y_{1} , \cdots ,y_{m}\right )$. 


The self-attention in conventional Transformer is calculated as following:
\begin{gather}
e_{ij}=\frac{Q\left ( x_{i} \right )K\left ( x_{j} \right )^{T}}{\sqrt{d_{k}}} \\
\alpha _{ij}=\frac{ \exp e_{ij}}{\sum_{l=1}^{n}\exp e_{il}}
\end{gather}
where $e_{ij}$ measures similarity between inputs, $\alpha _{ij}$ is the attention weight, $Q\left ( \cdot  \right )$ and $K\left ( \cdot  \right )$ are the projection functions from the input space to the query space and the key space, respectively, and $d_{k}$ represents the dimensions of the queries and keys. Then, the value is weighted by $\alpha _{ij}$ to calculate the hidden state $z_{i}$:
\begin{equation}
z_{i}=\sum_{j=1}^{n} \alpha _{ij}V\left ( x_{j} \right )
\end{equation}
where $V\left ( \cdot  \right )$ is a projection function from the input space to the value space. The final encoder output is a hidden states sequence $\mathbf{z}\in \mathbb{R}^{n\times d_{z}}$, where $d_{z}$ is the dimension of the hidden states. The per-layer complexity of self-attention is $O(n^{2}\cdot d)$ \cite{NIPS2017_7181}, where $n$ is the sequence length and $d$ is the representation dimension.

\begin{figure}[t]
\centering
\includegraphics[width=7.2cm]{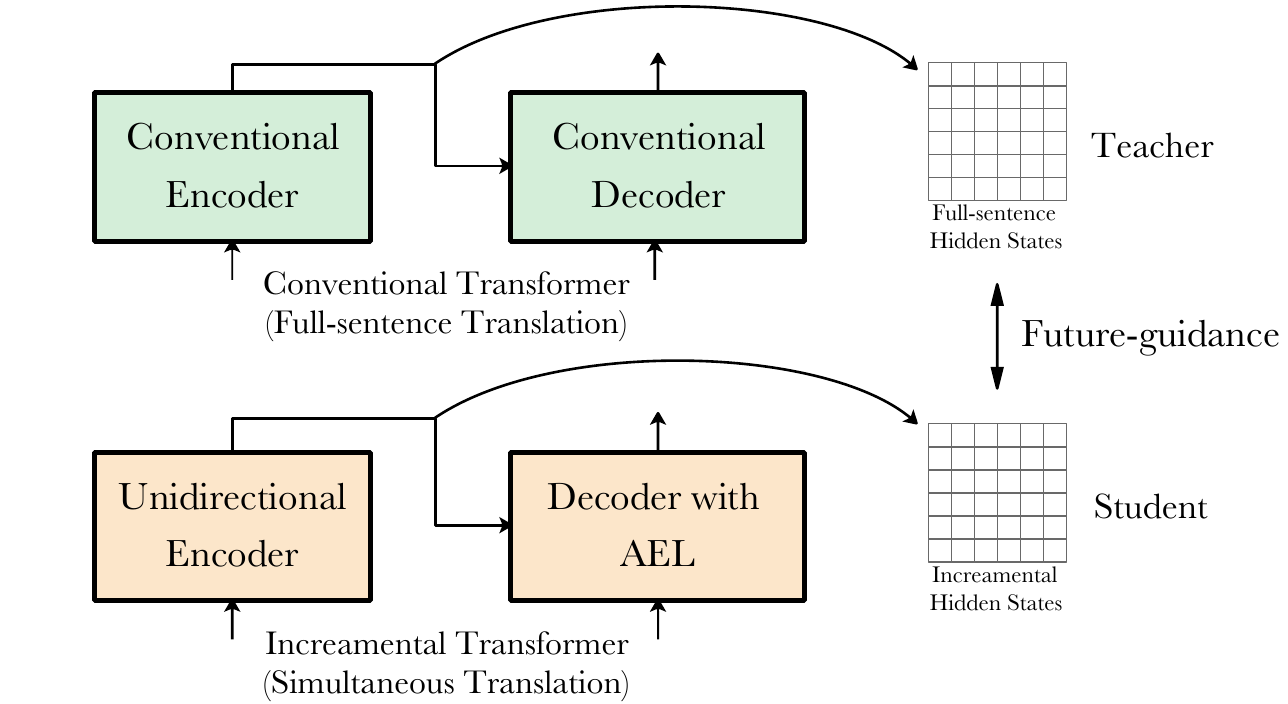}
\caption{The architecture of the proposed method. The lower part is the incremental Transformer, while the upper part is the conventional Transformer. A knowledge distillation is applied between the hidden states for future-guidance.}
\label{model2}
\end{figure}

\subsection{Wait-k Policy}
Wait-k policy \cite{ma-etal-2019-stacl} refers to waiting for $k$ source tokens first, and then reading and writing alternately, i.e., the output always delays $k$ tokens after the input. 


Define $g\left ( t \right )$ as a monotonic non-decreasing function of $t$, which represents the number of source tokens read in when outputting the target token $y_{t}$. For the wait-k policy, $g\left ( t \right )$ is calculated as:
\begin{equation}
g\left ( t \right )=\min\left \{ k+t-1, \left | \mathbf{x} \right | \right \},  t=1,2,\cdots
\end{equation}

To simulate ``prefix-to-prefix'' training, the source tokens participating in self-attention is limited to less than $g\left ( t \right )$:
\begin{gather}
e_{ij}^{\left ( t \right )}=\left\{\begin{matrix}\frac{Q\left ( x_{i} \right )K\left ( x_{j} \right )^{T}}{\sqrt{d_{k}}}
 & \mbox{if } i,j\leq  g\left ( t \right )\\ -\infty 
 & \mbox{otherwise}
\end{matrix}\right. \\
\alpha_{ij}^{\left ( t \right )}=\left\{\begin{matrix} \frac{ \exp e_{ij}^{(t)}}{\sum_{l=1}^{n}\exp e_{il}^{(t)}}
 & \mbox{if } i,j\leq  g\left ( t \right )\\ 0
 & \mbox{otherwise}
\end{matrix}\right.
\end{gather}
The hidden state of $i^{th}$ source token at decoding step $t$ is calculated as:
\begin{equation}
z_{i}^{(t)}=\sum_{j=1}^{n} \alpha _{ij}^{(t)}V\left ( x_{j} \right )
\end{equation}
The new hidden states is $\mathbf{z}^{\left (T  \right )}\in \mathbb{R}^{n\times d_{z}\times T}$, where $T$ represents the total number of decoding steps. Since the source token that read in changed at different decoding step, the hidden states sequence $\mathbf{z}^{t}$ at each step needs to be recalculated. The per-layer complexity of self-attention in wait-k policy is up to $O(n^{3}\cdot d)$, which greatly increase by $n$ times compared with full-sentence NMT.

\section{The Proposed Method}
\begin{figure}[t]
\centering
\includegraphics[width=7.2cm]{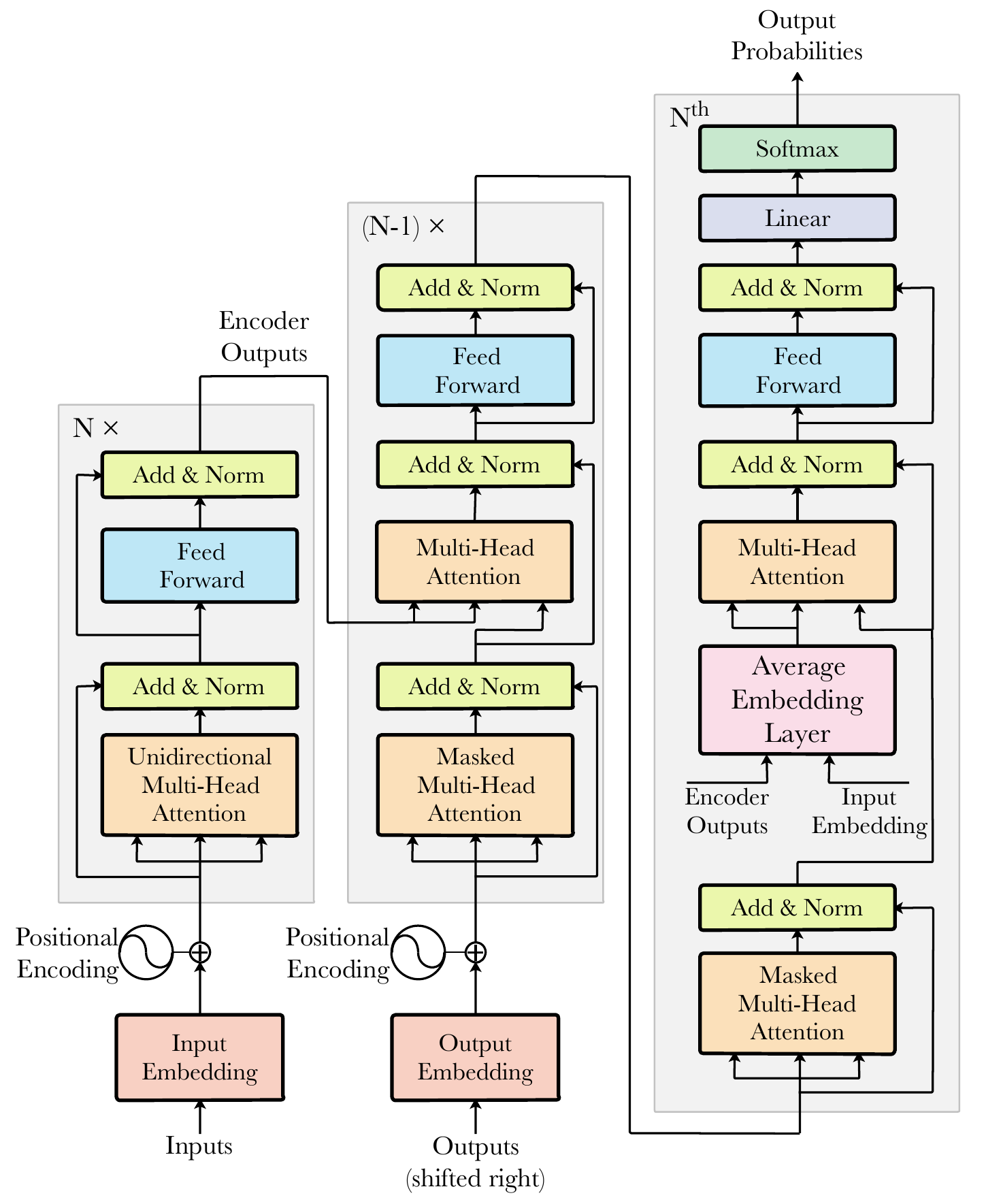}
\caption{The architecture of the proposed incremental Transformer with average embedding layer. The rightmost column represents the last layer of the decoder, including the average embedding layer}
\label{model}
\end{figure}

Our method is based on wait-k policy and consists of two components: incremental Transformer and conventional Transformer (full-sentence NMT). The architecture of the proposed method is shown in Figure \ref{model2}. Conventional Transformer is a standard Transformer \cite{NIPS2017_7181}, used as the teacher of incremental Transformer for knowledge distillation. Incremental Transformer is the proposed structure for ST, and the architecture of the incremental Transformer is shown in Figure \ref{model}. 

Incremental Transformer contains a unidirectional encoder (left-to-right) and a decoder with Average Embedding Layer (AEL). To avoid the recalculation of the source hidden states, we applied a unidirectional encoder, in which each token can only pay attention to the previous tokens. To establish the attention to the later tokens in the consumed source, an average embedding layer is added to the last layer of decoder, compensating for the lack of attention. The model can attend all consumed source through unidirectional encoder and AEL, without much more complexity. Specific details are introduced following.

\subsection{Incremental Transformer}
\subsubsection{Unidirectional Encoder}

Since the wait-k poliy with the bidirectional encoder take a high training complexity caused by recalculation, we apply a unidirectional encoder (left-to-right), where each source token can only focus on its previous tokens. The self-attention in unidirectional encoder is calculated as:
\begin{gather}
e_{ij}=\left\{\begin{matrix}\frac{Q\left ( x_{i} \right )K\left ( x_{j} \right )^{T}}{\sqrt{d_{k}}}
 & \mbox{if } j\leq i\leq  g\left ( t \right )\\ -\infty 
 & \mbox{otherwise}
\end{matrix}\right.\\
\alpha_{ij}=\left\{\begin{matrix} \frac{ \exp e_{ij}}{\sum_{l=1}^{n}\exp e_{il}}
 & \mbox{if } j\leq i\leq  g\left ( t \right )\\ 0
 & \mbox{otherwise}
\end{matrix}\right.
\end{gather}
Due to the characteristics of wait-k policy: $g\left ( t \right )=\min\left \{ k+t-1, \left | \mathbf{x} \right | \right \}$, $g\left ( t \right )$ changes linearly over the decoding step $t$. The calculation of $\alpha_{ij}$ can be decomposed into a unidirectional attention among all source tokens, and then mask out the part outside the $g\left ( t \right )$ through a mask matrix. 

\subsubsection{Decoder with AEL}

The unidirectional encoder only need to calculate the representation of the new source token, avoiding the complicated recalculation. But obviously, the price is that the front token lacks some attention to its later tokens. To make up for this, we propose an average embedding layer to summarize the information of all consumed sources. Since applying AEL in more decoder layers will gradually increase computational complexity, we only add AEL into the last layer of the decoder after trade-off between the computational complexity and translation quality.

\begin{figure}[t]
\centering
\includegraphics[width=7.5cm]{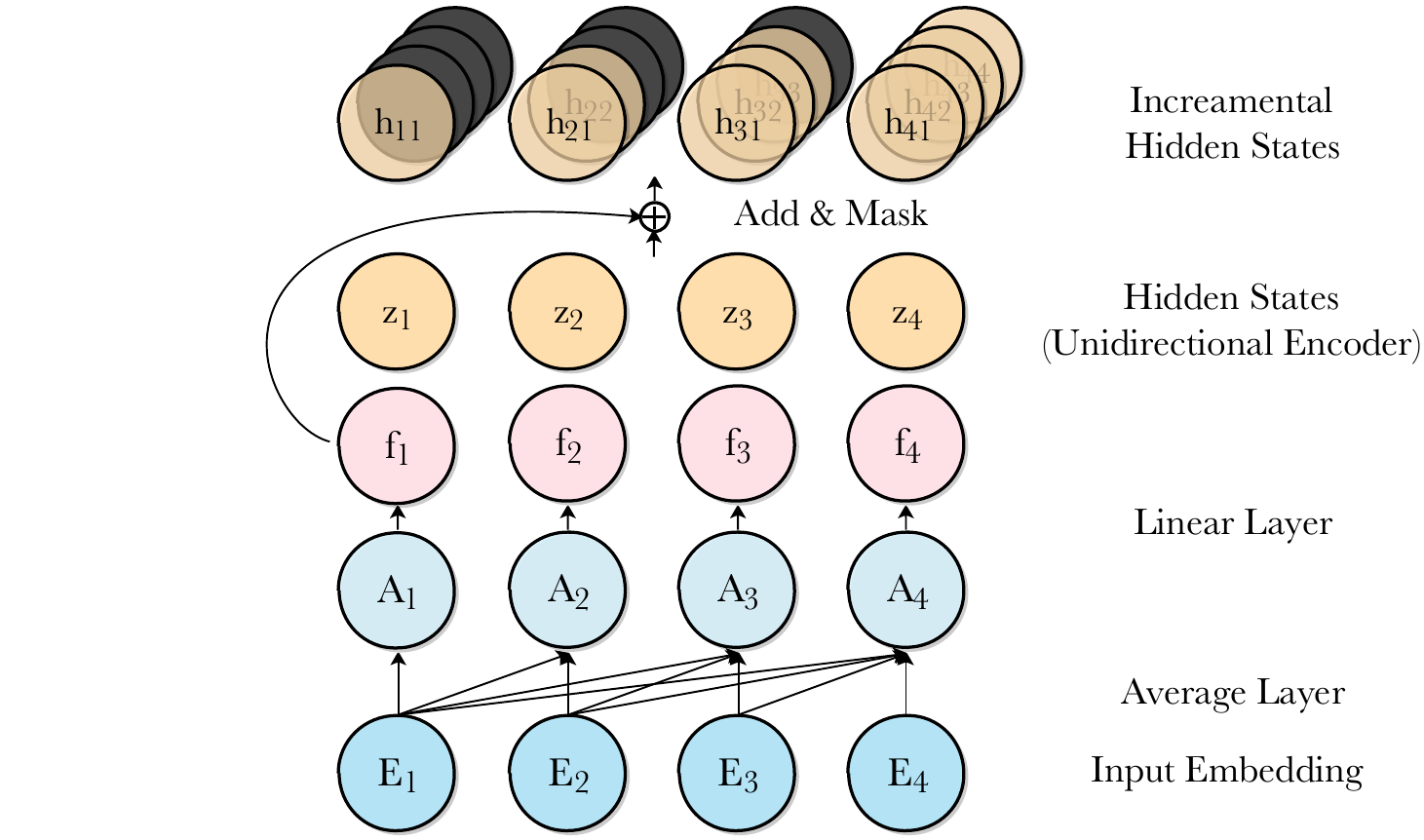}
\caption{The architecture of average embedding layer. For clarity, we show an example with only four tokens ($n=4$) and wait-2 policy ($k=2$).}
\label{ael}
\end{figure}

As shown in Figure \ref{ael}, through AEL, the average embedding of all consumed source is added into the unidirectional hidden states to focus on the later tokens. The inputs of the average embedding layer are hidden states $\mathbf{z}=\left ( z_{1} , \cdots ,z_{n}\right )$ and input embedding $\mathbf{E}=\left ( E_{1} , \cdots ,E_{n}\right )$. First, AEL performs an average operation on the input embedding:
\begin{equation}
A_{i}=\frac{1}{i}\sum_{j=1}^{i} E_{j}
\end{equation}
where $A_{i}\in \mathbb{R}^{d_{model}}$ is the average embedding of the first $i$ tokens. Since the average is not a complicated calculation, we can use the mask matrix to parallelize average operation.


To map $\mathbf{A}$ from the embedding space to the hidden states space, we applied a linear layer to get $\mathbf{f}$: 
\begin{equation}
 f_{i}=\mathbf{W}A_{i}
\end{equation}
where $\mathbf{W}\in \mathbb{R}^{d_{model}\times d_{model}}$ is a trainable parameter matrix, and $f_{i}$ represents the average information of the first $i$ tokens. Then, $\mathbf{f}$ is added to the hidden states of the tokens have been read in:
\begin{equation}
h_{ij}=\left\{\begin{matrix}
 f_{i}+z_{j}& j\leq i\\ 
 \mathbf{0}&\textrm{otherwise }
\end{matrix}\right.
\end{equation}
where $h_{ij}$ represents the new hidden state of the $j^{th}$ token when reading the first $i$ source tokens. Through AEL, the incremental hidden states is $\mathbf{h}\in \mathbb{R}^{n\times n\times d_{model}}$. 

Through unidirectional encoder and AEL, the incremental hidden states include the information of both previous tokens and later tokens. In the subsequent cross-attention, at the decoding step $t$, the decoder does multi-head attention with the slice $\mathbf{h}_{g(t)}$ in the incremental hidden state, where $g(t)$ the number of source tokens read in at $t$.

\subsection{Knowledge Distillation}
The most critical issue for ST is to achieve both high translation quality and low latency. With a guaranteed low latency, our method enables the model to predict future implicitly and capture some future source information that helps to determine sentence structure and translate.

As shown in the Figure \ref{model2}, We introduced a conventional Transformer as the teacher of the incremental Transformer, and shorten the distance between the hidden states of them. During training, the incremental Transformer encodes the incremental source, while the conventional Transformer can encode the complete source. Through knowledge distillation, conventional Transformer can teach the incremental Transformer to encode some future source information. For better distillation effect, we apply $L_{2}$ regularization term between the hidden states of them, where is closer to the source. The $L_{2}$ regularization term is calculated as:
\begin{equation}
\mathcal{L}\left ( \mathbf{z}^{incr} ,\mathbf{z}^{full}\right )=\frac{1}{n}\sum_{i=1}^{n}\left \|  z_{i}^{incr} -z_{i}^{full}\right \|^{2}   
\end{equation}
where $\mathbf{z}^{incr}$ and $\mathbf{z}^{full}$ represent the hidden states of incremental Transformer and conventional Transformer, respectively. 

Both incremental Transformer and conventional Transformer are trained with cross-entropy loss. The cross-entropy losses of incremental Transformer $\mathcal{L}\left ( \theta_{incr}  \right )$ and conventional Transformer $\mathcal{L}\left ( \theta_{full}  \right )$ on train data $D$ are respectively expressed as:
\begin{gather}
    \mathcal{L}\left ( \theta_{incr}  \right )=-\sum_{\left (\mathbf{x},\mathbf{y^{\star }}  \right )\in D}\log p_{incr}\left ( \mathbf{y}^{\star }\mid\left ( \mathbf{x},\theta_{incr}  \right )  \right )\\
    \mathcal{L}\left ( \theta_{full}  \right )=-\sum_{\left (\mathbf{x},\mathbf{y^{\star }}  \right )\in D}\log p_{full}\left ( \mathbf{y}^{\star }\mid\left ( \mathbf{x},\theta_{full}  \right )  \right )
\end{gather}
Then, the total loss $\mathcal{L}$ is calculated as:
\begin{equation}
    \mathcal{L}=\mathcal{L}\left ( \theta_{incr}  \right )+\mathcal{L}\left ( \theta_{full}  \right )+\lambda \mathcal{L}\left ( \mathbf{z}^{incr} ,\mathbf{z}^{full}\right ) \label{eq16}
\end{equation}
where $\lambda$ is an hyper-parameter controlling the importance of the penalty term, we set $\lambda=0.1$ in our experiments. We conducted experiment to compare the performance between pre-training a fixed conventional Transformer and jointly training the incremental Transformer and conventional Transformer in Table \ref{pre-train}, and finally apply jointly training them.

\section{Experiments}

\subsection{Datasets}

We conducted experiments on Chinese $\rightarrow $ English and German $\rightarrow $ English datasets.

\textbf{Chinese} $\rightarrow $ \textbf{English} (Zh-En) The training set consists of about 1.25M sentence pairs from LDC corpora\footnote{The corpora include LDC2002E18, LDC2003E07, LDC2003E14, Hansards portion of LDC2004T07, LDC2004T08 and LDC2005T06.}. We use MT02 as the validation set and MT03, MT04, MT05, MT06, MT08 as the test sets, each with 4 English references. We first tokenize and lowercase English sentences with the Moses\footnote{\url{http://www.statmt.org/moses/}}, and segmente the Chinese sentences with the Stanford Segmentor\footnote{\url{https://nlp.stanford.edu/}}. We apply BPE \cite{sennrich-etal-2016-neural} with 30K merge operations on all texts.

\textbf{German} $\rightarrow $ \textbf{English} (De-En) The training set consists of about 4.5M sentence pairs from WMT15 \footnote{\url{http://www.statmt.org/wmt15/translation-task.html}} De-En task. We use news-test2013(3000 sentence pairs) as the validation set and news-test2015(2169 sentence pairs) as the test set. We apply BPE with 32K merge operations, and the vocabulary is shared across languages.

\begin{table}[t]
\centering
\begin{tabular}{c||l|c|cc} \hline
                     &                & \textbf{Teacher} & \multicolumn{2}{c}{\textbf{Student}} \\ \hline
                     &                & BLEU    & AL           & BLEU          \\ \hline
\multirow{2}{*}{$k=9$} & Pre-training   & 45.13   & 9.81         & 40.57         \\
                     & Joint training & 44.91   & 9.63         & \textbf{41.86}         \\ \hline
\multirow{2}{*}{$k=7$} & Pre-training   & 45.13   & 7.81         & 39.71         \\
                     & Joint training & 44.88   & 8.11         & \textbf{40.73}         \\ \hline
\multirow{2}{*}{$k=5$} & Pre-training   & 45.13   & 6.50         & 38.39         \\
                     & Joint training & 44.84   & 6.26         & \textbf{40.00}         \\ \hline
\multirow{2}{*}{$k=3$} & Pre-training   & 45.13   & 4.62         & 37.00         \\
                     & Joint training & 44.62   & 4.43         & \textbf{38.28}         \\ \hline
\multirow{2}{*}{$k=1$} & Pre-training   & 45.13   & 2.34         & 32.11         \\
                     & Joint training & 44.58   & 2.32         & \textbf{34.20}         \\ \hline
\end{tabular}
\caption{Comparison between pre-training a fixed conventional Transformer and jointly training incremental Transformer (Student) and conventional Transformer (Teacher), testing on Zh-En validation set. We show the performance of the final teacher model and student model. Note that the teacher model is evaluated on full-sentence NMT.}
\label{pre-train}
\end{table}

\begin{table*}[t]
\centering
\begin{tabular}{l||l|ccccc|cc|c|c}

\hline
                              & \multirow{2}{*}{} & \textbf{MT03}  & \textbf{MT04}  & \textbf{MT05}  & \textbf{MT06}  & \textbf{MT08}  & \multicolumn{2}{c|}{\textbf{AVERAGE}} & \multirow{2}{*}{\textbf{$\mathbf{\Delta}$}} & \multirow{2}{*}{\textbf{\begin{tabular}[c]{@{}c@{}}Training Time\\      (secs/b)\end{tabular}}} \\
                              &                   & \multicolumn{5}{c|}{BLEU}                                                           & AL          & BLEU                &                                    &                                                                                                 \\ \hline
\multirow{2}{*}{offline}                              & bi-transformer    & 44.56          & 45.69          & 45.28          & 44.63          & 34.51          & 28.83       & 42.93               &                                    & 0.31                                                                                            \\
                              & uni-transformer   & 43.22          & 44.40          & 43.12          & 42.31          & 32.51          & 28.82       & 41.11               &                                    & 0.31                                                                                            \\ \hline
\multirow{5}{*}{$k=9$} & baseline(bi)      & 40.35          & 42.21          & 40.21          & 40.78          & 32.45          & 9.99        & 39.20               &                                    &                                   9.92                                                              \\
                              & baseline(uni)     & 39.42          & 42.08          & 40.33          & 40.12          & 31.59          & 9.99        & 38.71               &                                    &         0.31                                                                                        \\
                              & $\;\;$+AEL             & 40.77          & 42.27          & 40.11          & 40.77          & 32.17          & 10.09       & 39.22               & +0.51                              & 0.41                                                                                            \\
                              & $\;\;$+Teacher          & 41.52          & \textbf{43.05}          & \textbf{41.75}          & 41.59          & \textbf{33.12}          & 9.74        & 40.21               & +0.99                              & 0.78                                                                                            \\
                              & $\;\;$+AEL+Teacher      &  \textbf{41.75}              &   43.03             &  41.63              &   \textbf{41.76}             &   33.06             &  9.73           &    \textbf{40.25}                 &   \textbf{+1.54}                                 & 0.80                                                                                            \\ \hline
\multirow{5}{*}{$k=7$} & baseline(bi)      & 40.27          & 41.94          & 39.90          & 40.35          & 31.84          & 8.05        & 38.86               &                                    &                                   10.26$\;\;$                                                              \\
                              & baseline(uni)     & 38.79          & 41.12          & 38.77          & 39.13          & 30.61          & 8.01        & 37.68               &                                    &          0.31                                                                                       \\
                              & $\;\;$+AEL             & 39.81          & 41.66          & 38.81          & 40.14          & 31.16          & 8.17        & 38.32               & +0.63                              & 0.41                                                                                            \\
                              & $\;\;$+Teacher          & \textbf{40.51}          & 41.81          & \textbf{40.35}          & \textbf{40.90}          & 32.16          & 8.31        & 39.15               & +1.46                              & 0.79                                                                                            \\
                              & $\;\;$+AEL+Teacher      &   40.41             &    \textbf{42.08}            &   40.29             &   40.44             &  \textbf{32.94}              &   8.10          &    \textbf{39.23}                 &     \textbf{+1.55}                               & 0.81                                                                                            \\ \hline
\multirow{5}{*}{$k=5$} & baseline(bi)      & 40.12          & 41.46          & 39.58          & 40.21          & \textbf{31.57} & 6.34        & 38.59               &                                    &                                   10.70$\;\;$                                                               \\
                              & baseline(uni)     & 37.09          & 39.62          & 37.78          & 37.66          & 29.82          & 6.27        & 36.39               &                                    &          0.31                                                                                       \\
                              & $\;\;$+AEL             & 38.74          & 40.11          & 38.36          & 39.04          & 30.30          & 6.06        & 37.31               & +0.92                              & 0.41                                                                                            \\
                              & $\;\;$+Teacher          & 39.47          & 40.42          & 38.82          & 39.78          & 30.05          & 6.24        & 37.71               & +1.31                              & 0.82                                                                                            \\
                              & $\;\;$+AEL+Teacher      & \textbf{40.15} & \textbf{41.53} & \textbf{39.58} & \textbf{40.59} & 31.29          & 5.98        & \textbf{38.63}      & \textbf{+2.23}                              & 0.83                                                                                            \\ \hline
\multirow{5}{*}{$k=3$} & baseline(bi)      & 37.08          & \textbf{39.11} & 36.69          & 37.20          & 28.28          & 4.15        & 35.67               &                                    &                                   11.11$\;\;$                                                               \\
                              & baseline(uni)     & 35.94          & 36.98          & 34.64          & 34.80          & 26.48          & 4.42        & 33.77               &                                    &          0.31                                                                                       \\
                              & $\;\;$+AEL             & 37.40          & 38.72          & 36.64          & 36.59          & 28.06          & 4.11        & 35.48               & +1.71                              & 0.41                                                                                            \\
                              & $\;\;$+Teacher          & 37.42          & 38.94          & 37.13          & 37.37          & \textbf{29.58} & 4.53        & 36.09               & +2.32                              & 0.84                                                                                            \\
                              & $\;\;$+AEL+Teacher      & \textbf{38.15} & 38.88          & \textbf{37.14} & \textbf{37.46} & 28.98          & 4.41        & \textbf{36.12}      & \textbf{+2.35}                              & 0.86                                                                                            \\ \hline
\multirow{5}{*}{$k=1$} & baseline(bi)      & 32.67          & 34.51          & 32.55          & 32.04          & 24.79          & 2.45        & 31.31               &                                    &                                   15.11$\;\;$                                                               \\
                              & baseline(uni)     & 31.99          & 33.75          & 31.47          & 31.56          & 23.86          & 2.71        & 30.53               &                                    &          0.31                                                                                       \\
                              & $\;\;$+AEL             & 32.97          & 34.41          & 32.37          & 32.04          & 24.16          & 2.29        & 31.19               & +0.66                              & 0.41                                                                                            \\
                              & $\;\;$+Teacher          & 33.95          & 34.51          & 33.07          & 33.17          & 25.14          & 2.35        & 31.97               & +1.44                              & 0.84                                                                                            \\
                              & $\;\;$+AEL+Teacher      & \textbf{34.21} & \textbf{35.10} & \textbf{33.11} & \textbf{33.72} & \textbf{25.19} & 2.37        & \textbf{32.27}      & \textbf{+1.74}                              & 0.86                                                                                            \\ \hline
\end{tabular}
\caption{Translation quality (4-gram BLEU), latency (AL), and training speed (seconds/batch) on Zh-En simultaneous translation. Since our proposed method and baseline belong to the fixed policy, there is almost no difference in latency. Therefore, we display the results in the form of table to highlight the details of the improvement in translation quality and training speed.}
\label{zh-en result}
\end{table*}

\subsection{Systems Setting}
We conducted experiments on the following systems:


{\bf bi-Transformer}: offline model. Full-sentence NMT based on Transformer with bidirectional encoder.

{\bf uni-Transformer}: offline model. Full-sentence NMT based on Transformer with unidirectional encoder.

{\bf baseline(bi)}: wait-k policy based on Transformer with bidirectional encoder \cite{ma-etal-2019-stacl}.

{\bf baseline(uni)}: wait-k policy based on Transformer with unidirectional encoder.

{\bf +Teacher}: only add a conventional Transformer as the teacher model based on Transformer with unidirectional encoder. The encoder of teacher model is bidirectional.

{\bf +AEL}: only add average embedding layer we proposed based on Transformer with unidirectional encoder.

{\bf +AEL+Teacher}: add both AEL and teacher model based on Transformer with unidirectional encoder.

The implementation of our method is adapted from Fairseq Library \cite{ott-etal-2019-fairseq}. The parameters of the incremental Transformer we proposed are exactly the same as the standard wait-k \cite{ma-etal-2019-stacl}, while the conventional Transformer is the same as the original Transformer \cite{NIPS2017_7181}. 

\subsection{Comparison between Joint Training and Pre-training}
\label{section:my}

Before the main experiment, we compared the performance of `+Teacher' between pre-training a fixed conventional Transformer or jointly training incremental Transformer and conventional Transformer on Zh-En validation set. 

As shown in Table \ref{pre-train}, jointly training makes the model get better performance than pre-training. The reason is that the teacher model is for full-sentence MT, while the student model is for ST, and the two have inherent differences in the hidden states distribution. Since the decoding policy is incremental at the inference time, we should not let the incremental Transformer learn from the conventional Transformer without any difference, but narrow the distance between them, helping the student model maintain the characteristics of incremental decoding. Similarly, \cite{dalvi-etal-2018-incremental,ma-etal-2019-stacl} pointed out that if the full-sentence NMT model is directly used for ST, the translation quality will be significantly reduced. Besides, with joint-training, the performance of the final teacher model will not be greatly affected, which can still guide the student model. Therefore, we jointly train the incremental Transformer and conventional Transformer with the loss in Eq.(\ref{eq16}).

\subsection{Comparison with baseline}

We set standard wait-k policy as the baseline and compare with it. For evaluation metric, we use BLEU \cite{papineni-etal-2002-bleu} and AL\footnote{The calculation of AL is as \url{https://github.com/SimulTrans-demo/STACL}.} \cite{ma-etal-2019-stacl}  to measure translation quality and latency, respectively.
Table \ref{zh-en result} reports translation quality (BLEU), latency (AL) and training time of our method, baseline and offline model on Zh-En simultaneous translation, and `AVERAGE' is average on all test sets. Table \ref{de-en result} reports the result on De-En simultaneous translation.

We first notice that the training speed of the baseline(bi) is too slow, where the training time of each batch is about 36.84 times (average on different $k$) that of the offline model. As $k$ decreases, the training time will gradually increase, until $k=1$, the training time even increase by 48.74 times. When $k$ is smaller, the number of tokens waiting at the beginning is less, and the number of recalculation of encoder hidden states increases rapidly. After adopting AEL in Transformer with unidirectional encoder, our method avoids the recalculation of encoder hidden states and also makes up for the lack of attention of the unidirectional encoder. The training speed of `+AEL' is about 27.86 times (average on different $k$) faster than that of baseline(bi), while the translation quality is equivalent to that of baseline(bi).

After adding the conventional Transformer to guide incremental Transformer, `+Teacher' improved about 1.5 BLEU (average on different $k$) over the baseline(uni). Note that in the case of low latency (smaller $k$), our method improves especially. When $k$ is very small, the model waits for a very few tokens, so that the prediction of the future is more important at a low latency. In general, after applying AEL and Teacher model, the training speed of `+AEL+Teacher' is increased by about 13.67 times, and translation quality improves about 1.88 BLEU on Zh-En and 0.91 BLEU on De-En (average on different $k$).

For the case of different waiting time $k$ between training and testing, \cite {ma-etal-2019-stacl} pointed out that the best results when testing with wait-$j$ policy are often from a model trained with a larger wait-$i$ policy (where $i>j$),
which shows that the model trained with more source information performs better. Table \ref{diff k} shows the results of the proposed method using wait-$i$ policy during training and testing with wait-$j$ policy. The best results are basically obtained when $j=i$, since future-guided methods inspires incremental Transformer learn implicit future information. It is worth mentioning that the best result for wait-1 testing still comes from wait-7 training model. We presume the reason is that although wait-1 model learns some future information, the delay of one token still contains too little information.
\begin{table}[t]
\centering
\begin{tabular}{c||l|cc|c}
\hline
                      &                 & \textbf{AL}              & \textbf{BLEU}              & $\mathbf{\Delta}$             \\ \hline
\multirow{2}{*}{offline} & bi-Transformer  & \multicolumn{1}{c}{28.60$\;\;$} & \multicolumn{1}{c|}{31.42} & \multicolumn{1}{l}{} \\
 & uni-Transformer & \multicolumn{1}{c}{28.70$\;\;$} & \multicolumn{1}{c|}{30.12} & \multicolumn{1}{l}{} \\ \hline
\multirow{3}{*}{$k=9$}  & baseline(bi)    & 9.36                     & 28.48                      &                      \\
                      & baseline(uni)   & 9.24                     & 28.10                      &                      \\
                      & $\;\;$+AEL+Teacher    & 9.25                     & \textbf{29.42}                      & +1.32                \\ \hline
\multirow{3}{*}{$k=7$}  & baseline(bi)    & 7.44                     & 28.09                      &                      \\
                      & baseline(uni)   & 7.83                     & 27.84                      &                      \\
                      & $\;\;$+AEL+Teacher    & 7.90                     & \textbf{28.38}                      & +0.54                \\ \hline
\multirow{3}{*}{$k=5$}  & baseline(bi)    & 5.58                     & 26.38                      &                      \\
                      & baseline(uni)   & 5.78                     & 25.73                      &                      \\
                      & $\;\;$+AEL+Teacher    & 5.74                     & \textbf{26.97}                      & +1.24                \\ \hline
\multirow{3}{*}{$k=3$}  & baseline(bi)    & 3.48                     & 24.18                      &                      \\
                      & baseline(uni)   & 3.91                     & 24.04                      &                      \\
                      & $\;\;$+AEL+Teacher    & 3.95                     & \textbf{24.39}                      & +0.35                \\ \hline
\multirow{3}{*}{$k=1$}  & baseline(bi)    & 1.60                     & 18.48                      &                      \\
                      & baseline(uni)   & 1.32                     & 18.29                      &                      \\
                      & $\;\;$+AEL+Teacher    & 1.31                     & \textbf{19.36}                      & +1.07                \\ \hline
\end{tabular}
\caption{Translation quality (BLEU) and latency (AL) on De-En simultaneous translation.}
\label{de-en result}
\end{table}

\begin{table}[t]
\centering
\setlength{\tabcolsep}{1.1mm}
\begin{tabular}{|c|c|c|c|c|c|}
\hline
\diagbox{Train $k$}{Test $k$}  & 1              & 3              & 5              & 7              & 9              \\ \hline
1 & 32.27          & 35.07          & 35.95          & 36.17          & 35.77          \\ \hline
3 & 32.65 & \textbf{36.12} & 38.05          & 38.70          & 39.99          \\ \hline
5 & 31.95          & 35.35          & \textbf{38.63} & 38.62          & 39.28          \\ \hline
7 & \textbf{32.74}          & 36.04          & 38.37          & \textbf{39.23} & 39.64          \\ \hline
9 & 31.91          & 35.49          & 37.91          & 38.99          & \textbf{40.25} \\ \hline
\end{tabular}
\caption{Results(average BLEU over all Zh-En test sets) of proposed method `+AEL+Teacher' using wait-$i$ policy during training and wait-$j$ policy during testing.}
\label{diff k}
\end{table}

\begin{figure}[t]
  \centering
  \subfigure[without $L_{2}$ regularization\label{tsne-a}]{\includegraphics[width=1.5in]{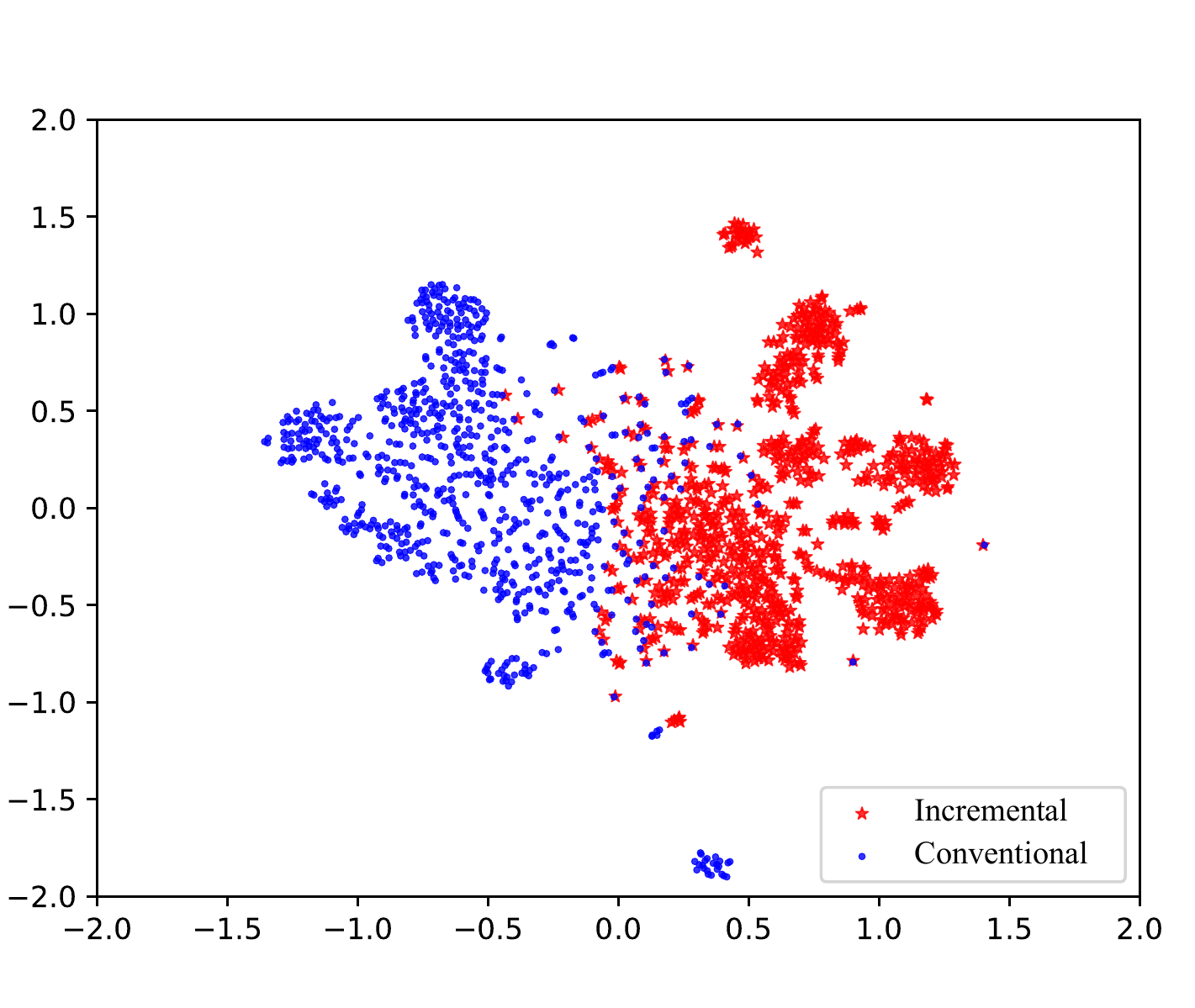}}
  \subfigure[with $L_{2}$ regularization\label{tsne-b}]{\includegraphics[width=1.5in]{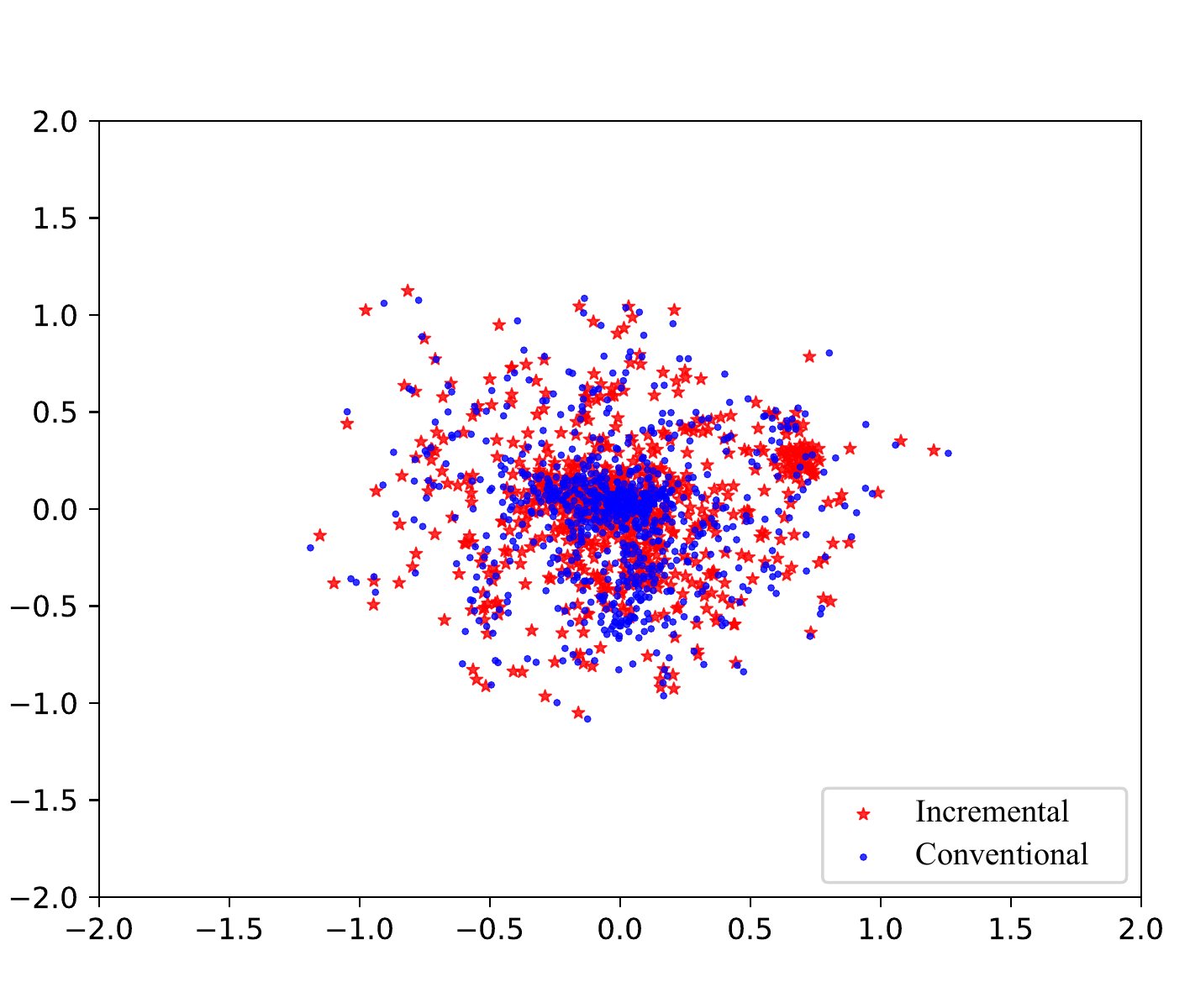}}
  \caption{The distribution of the hidden states of incremental Transformer and conventional Transformer on the Zh-En validation set. Red stars represents the hidden states of the incremental Transformer, while the blue dots represents the hidden states of the conventional Transformer.} 
  \label{tsne}
\end{figure}

\subsection{Impact of the Knowledge Distillation}

\begin{table*}[t]
\setlength{\tabcolsep}{1.5mm}
\centering
\begin{tabular}{c|c|c|c|c|c|c|c|c|c|c}
\hline
              & \multicolumn{2}{c|}{$k=1$} & \multicolumn{2}{c|}{$k=3$} & \multicolumn{2}{c|}{$k=5$} & \multicolumn{2}{c|}{$k=7$} & \multicolumn{2}{c}{$k=9$} \\ \cline{2-11} 
              & baseline  & +Teacher & baseline  & +Teacher & baseline  & +Teacher & baseline  & +Teacher & baseline  & +Teacher \\ \hline
\textbf{Absent} & 54.88              & \textbf{59.82}        & 61.34              & \textbf{63.26}        & 63.54              & \textbf{65.38}        & 70.72              & \textbf{71.80}        & 70.48              & \textbf{71.57}        \\ \hline
\textbf{Present}  & 82.47              & \textbf{83.32}        & 84.76              & \textbf{85.22}        & 85.33              & \textbf{86.04}        & 85.94              & \textbf{86.51}        & 86.25              & \textbf{86.92}        \\ \hline
\end{tabular}
\caption{1-gram score of baseline and `+Teacher' on Absent. set and Present. set, respectively. `+Teacher' indicates applying a conventional Transformer for future-guidance. `Absent' represents the aligned source token has not been read in when generating the target token, `Present' represents the aligned source token has been read in when generating the target token.}
\label{pred}
\end{table*}
\begin{table}[]
\centering
\begin{tabular}{l||l|c|c}
\hline
                     &                                                      & \textbf{AVG BLEU}             & \textbf{$\mathbf{\Delta}$} \\ \hline
\multirow{3}{*}{$k=9$} & baseline(uni)                                        & 38.71                           &                   \\
                     & $\;\;\;$+uni-Teacher & 39.72                           & +1.01             \\
                     & $\;\;\;$+bi-Teacher      & \textbf{40.21} & +1.50             \\ \hline
\multirow{3}{*}{$k=7$} & baseline(uni)                                        & 37.68                           &                   \\
                     & $\;\;\;$+uni-Teacher & 38.95                           & +1.27             \\
                     & $\;\;\;$+bi-Teacher      & \textbf{39.15} & +1.46             \\ \hline
\multirow{3}{*}{$k=5$} & baseline(uni)                                        & 36.39                           &                   \\
                     & $\;\;\;$+uni-Teacher & 37.50                           & +1.11             \\
                     & $\;\;\;$+bi-Teacher      & \textbf{37.71} & +1.32             \\ \hline
\multirow{3}{*}{$k=3$} & baseline(uni)                                        & 33.77                           &                   \\
                     & $\;\;\;$+uni-Teacher & 36.02                           & +2.25             \\
                     & $\;\;\;$+bi-Teacher      & \textbf{36.09} & +2.32             \\ \hline
\multirow{3}{*}{$k=1$} & baseline(uni)                                        & 30.53                           &                   \\
                     & $\;\;\;$+uni-Teacher & \textbf{32.08} & +1.55             \\
                     & $\;\;\;$+bi-Teacher      & 31.97                           & +1.44             \\ \hline
\end{tabular}
\caption{Comparison between the teacher model using unidirectional / bidirectional encoder, test on Zh-En test set. `+uni-Teacher' indicates using unidirectional encoder, while `+bi-Teacher' indicates using bidirectional encoder.}
\label{teacher}
\end{table}

Our method applies knowledge distillation with a $L_{2}$ regularization term. We reduce the dimension of the hidden states with t-Distributed Stochastic Neighbor Embedding (t-SNE) technique, and show the distribution in Figure \ref{tsne}. 
With the $L_{2}$ regularization term, the hidden states are fused with each other, which shows the source information extracted by incremental Transformer and conventional Transformer is more closer. Therefore, $L_{2}$ regularization term successfully makes incremental Transformer learn some future information from conventional Transformer.

Besides, to ensure that most of the improvement brought by `+Teacher' comes from the knowledge distillation between the full-sentence / incremental encoder, not due to the knowledge distillation between bidirectional / unidirectional encoder, we report the results of using teacher model with unidirectional / bidirectional encoder in Table \ref{teacher}. When using Transformer with unidirectional encoder as the teacher, our method can be improved about 1.5 BLEU. When the unidirectional encoder was replaced by the bidirectional encoder, the translation quality was only slightly further improved about 0.2 BLEU. When both the teacher model and the student model use unidirectional encoder, the improvement brought by knowledge distillation is still obvious, which shows that most of the improvement brought by our proposed method `+Teacher' comes from the knowledge distillation between the conventional Transformer (full-sentence) and the incremental Transformer.

\subsection{Prediction Accuracy}

To verify that our method implicitly embeds some future information through knowledge distillation, we tested the token prediction accuracy of `+Teacher' and baseline(bi) on Zh-En validation set. We first use GIZA++\footnote{\url{https://github.com/moses-ST/giza-pp.git}} to align the tokens between the generated translation and the source sentence. 
As a result, the $i^{th}$ target token is aligned with the $j^{th}$ source token. 
All the generated target tokens are divided into two sets: \textit{Present} and \textit{Absent}. If $j\leq \min\left (i+k-1,n  \right )$, the aligned source token of the $i^{th}$ generated token has been read when generating, thus the generated token belongs to \textit{Present} set. In contrast, if $j>  \min\left (i+k-1,n  \right )$, the aligned source token has not been read when generating, thus the generated token belongs to \textit{Absent} set, i.e., the generated target token is implicitly predicted by the model. Finally, the 1-gram score is calculated on both sets. 

The results are shown in Table \ref{pred}. After applying future-guidance with the teacher model, the token prediction accuracy improves. Our method improves more obviously when $k$ is smaller, since the small $k$ greatly limits the information that the model can read in. When $k$ is small, the predictive ability from data-driven becomes unreliable, and it is especially important to explicitly introduce the future-guidance. In addition, the accuracy on the Present. set does not decrease and improves slightly.

\section{Related Work}

The current research of ST is mainly divided into: precise read / write policy and stronger predictive ability.

For read / write policy, earlier methods were based on segmented translation \cite{bangalore-etal-2012-real,Cho2016,siahbani-etal-2018-simultaneous}. \citet {gu-etal-2017-learning} used reinforcement learning to train an agent to decide read / write. Recently, \citet {dalvi-etal-2018-incremental} proposed STATIC-RW, first performing $S$'s READs, then alternately performing $RW$'s WRITEs and READs. \citet {ma-etal-2019-stacl} proposed a wait-k policy, wherein begin synchronizing output after reading $k$ tokens. \citet{Zheng2019b} trained an agent by the input sentences and gold read / write sequence generated by rules. \citet {Zheng2019a} introduces a ``delay'' token $\left \{ \varepsilon  \right \}$ into the target vocabulary, and introduced limited dynamic prediction. \citet {Arivazhagan2019} proposed MILK, which uses a variable based on Bernoulli distribution to determine whether to output. \citet {Ma2019a} proposed MMA, the implementation of MILK based on Transformer.

Most of the previous methods use the unidirectional encoder \cite{Arivazhagan2019,Ma2019a} or fune-tuning a trained model \cite{dalvi-etal-2018-incremental} to reduce the computational cost. We proposed AEL to compensate for the lack of attention caused by unidirectional encoder.

For predicting future, \citet {Shigeki2000} applied pattern recognition to predict verbs in advance. \citet {grissom-ii-etal-2014-dont} used a Markov chain to predict the next word and final verb to eliminate delay bottlenecks between different word orders. \cite{oda-etal-2015-syntax} predict unseen syntactic constituents to help generate complete parse trees and perform syntax-based simultaneous translation. \citet {Alinejad2019} added a Predict operation to the agent based on \citet {gu-etal-2017-learning}, predicting the next word as an additional input. However, most of previous methods predict a specific word through a language model, while directly predicting specific words is prone to large errors which will cause mistakes in subsequent translations. Unlike the previous method, our method attempt to implicitly embed some future information in the model through future-guidance, avoiding the impact of inaccurate predictions.

\section{Conclusion}

In order to accelerate the training speed of the wait-k policy and use future information to guide the training, we propose future-guided incremental Transformer for simultaneous translation. With incremental Transformer and AEL, our method greatly accelerates the training speed about 28 times, meanwhile attends to all consumed source tokens. With future-guided training, the incremental Transformer successfully embeds some implicit future information and has a stronger predictive ability, without adding any latency or parameters in the inference time. Experiments show the proposed method outperform the baseline and achieve better performance on both training speed and translation quality.

\section{Acknowledgements}
We thank all the anonymous reviewers for their insightful and valuable comments. This work was supported by National Key R\&D Program of China (NO. 2018YFC0825201 and NO. 2017YFE0192900 ).

\bibliography{ref.bib}

\begin{thebibliography}{23}
\providecommand{\natexlab}[1]{#1}
\providecommand{\url}[1]{\texttt{#1}}
\providecommand{\urlprefix}{URL }
\expandafter\ifx\csname urlstyle\endcsname\relax
  \providecommand{\doi}[1]{doi:\discretionary{}{}{}#1}\else
  \providecommand{\doi}{doi:\discretionary{}{}{}\begingroup
  \urlstyle{rm}\Url}\fi

\bibitem[{Alinejad, Siahbani, and Sarkar(2018)}]{Alinejad2019}
Alinejad, A.; Siahbani, M.; and Sarkar, A. 2018.
\newblock Prediction Improves Simultaneous Neural Machine Translation.
\newblock In \emph{Proceedings of the 2018 Conference on Empirical Methods in
  Natural Language Processing}, 3022--3027. Brussels, Belgium: Association for
  Computational Linguistics.
\newblock \doi{10.18653/v1/D18-1337}.
\newblock \urlprefix\url{https://www.aclweb.org/anthology/D18-1337}.

\bibitem[{Arivazhagan et~al.(2019)Arivazhagan, Cherry, Macherey, Chiu, Yavuz,
  Pang, Li, and Raffel}]{Arivazhagan2019}
Arivazhagan, N.; Cherry, C.; Macherey, W.; Chiu, C.-c.; Yavuz, S.; Pang, R.;
  Li, W.; and Raffel, C. 2019.
\newblock {Monotonic Infinite Lookback Attention for Simultaneous Machine
  Translation}.
\newblock 1313--1323.
\newblock \doi{10.18653/v1/p19-1126}.

\bibitem[{Bangalore et~al.(2012)Bangalore, Rangarajan~Sridhar, Kolan, Golipour,
  and Jimenez}]{bangalore-etal-2012-real}
Bangalore, S.; Rangarajan~Sridhar, V.~K.; Kolan, P.; Golipour, L.; and Jimenez,
  A. 2012.
\newblock Real-time Incremental Speech-to-Speech Translation of Dialogs.
\newblock In \emph{Proceedings of the 2012 Conference of the North {A}merican
  Chapter of the Association for Computational Linguistics: Human Language
  Technologies}, 437--445. Montr{\'e}al, Canada: Association for Computational
  Linguistics.
\newblock \urlprefix\url{https://www.aclweb.org/anthology/N12-1048}.

\bibitem[{Chen et~al.(2020)Chen, Li, Jiang, Chen, and Liu}]{chen2020general}
Chen, Y.; Li, L.; Jiang, X.; Chen, X.; and Liu, Q. 2020.
\newblock A General Framework for Adaptation of Neural Machine Translation to
  Simultaneous Translation.

\bibitem[{Cho and Esipova(2016)}]{Cho2016}
Cho, K.; and Esipova, M. 2016.
\newblock {Can neural machine translation do simultaneous translation?}
  \urlprefix\url{http://arxiv.org/abs/1606.02012}.

\bibitem[{Dalvi et~al.(2018)Dalvi, Durrani, Sajjad, and
  Vogel}]{dalvi-etal-2018-incremental}
Dalvi, F.; Durrani, N.; Sajjad, H.; and Vogel, S. 2018.
\newblock Incremental Decoding and Training Methods for Simultaneous
  Translation in Neural Machine Translation.
\newblock In \emph{Proceedings of the 2018 Conference of the North {A}merican
  Chapter of the Association for Computational Linguistics: Human Language
  Technologies, Volume 2 (Short Papers)}, 493--499. New Orleans, Louisiana:
  Association for Computational Linguistics.
\newblock \doi{10.18653/v1/N18-2079}.
\newblock \urlprefix\url{https://www.aclweb.org/anthology/N18-2079}.

\bibitem[{Grissom~II et~al.(2014)Grissom~II, He, Boyd-Graber, Morgan, and
  Daum{\'e}~III}]{grissom-ii-etal-2014-dont}
Grissom~II, A.; He, H.; Boyd-Graber, J.; Morgan, J.; and Daum{\'e}~III, H.
  2014.
\newblock Don{'}t Until the Final Verb Wait: Reinforcement Learning for
  Simultaneous Machine Translation.
\newblock In \emph{Proceedings of the 2014 Conference on Empirical Methods in
  Natural Language Processing ({EMNLP})}, 1342--1352. Doha, Qatar: Association
  for Computational Linguistics.
\newblock \doi{10.3115/v1/D14-1140}.
\newblock \urlprefix\url{https://www.aclweb.org/anthology/D14-1140}.

\bibitem[{Gu et~al.(2017)Gu, Neubig, Cho, and Li}]{gu-etal-2017-learning}
Gu, J.; Neubig, G.; Cho, K.; and Li, V.~O. 2017.
\newblock Learning to Translate in Real-time with Neural Machine Translation.
\newblock In \emph{Proceedings of the 15th Conference of the {E}uropean Chapter
  of the Association for Computational Linguistics: Volume 1, Long Papers},
  1053--1062. Valencia, Spain: Association for Computational Linguistics.
\newblock \urlprefix\url{https://www.aclweb.org/anthology/E17-1099}.

\bibitem[{Hinton, Vinyals, and Dean(2015)}]{kn}
Hinton, G.; Vinyals, O.; and Dean, J. 2015.
\newblock Distilling the Knowledge in a Neural Network.

\bibitem[{Ma et~al.(2019)Ma, Huang, Xiong, Zheng, Liu, Zheng, Zhang, He, Liu,
  Li, Wu, and Wang}]{ma-etal-2019-stacl}
Ma, M.; Huang, L.; Xiong, H.; Zheng, R.; Liu, K.; Zheng, B.; Zhang, C.; He, Z.;
  Liu, H.; Li, X.; Wu, H.; and Wang, H. 2019.
\newblock {STACL}: Simultaneous Translation with Implicit Anticipation and
  Controllable Latency using Prefix-to-Prefix Framework.
\newblock In \emph{Proceedings of the 57th Annual Meeting of the Association
  for Computational Linguistics}, 3025--3036. Florence, Italy: Association for
  Computational Linguistics.
\newblock \doi{10.18653/v1/P19-1289}.
\newblock \urlprefix\url{https://www.aclweb.org/anthology/P19-1289}.

\bibitem[{Ma et~al.(2020)Ma, Pino, Cross, Puzon, and Gu}]{Ma2019a}
Ma, X.; Pino, J.~M.; Cross, J.; Puzon, L.; and Gu, J. 2020.
\newblock Monotonic Multihead Attention.
\newblock In \emph{International Conference on Learning Representations}.
\newblock \urlprefix\url{https://openreview.net/forum?id=Hyg96gBKPS}.

\bibitem[{Matsubara et~al.(2000)Matsubara, Kawaguchi, Toyama, and
  Inagaki}]{Shigeki2000}
Matsubara, Shigeki~Iwashima, K.; Kawaguchi, N.; Toyama, K.; and Inagaki, Y.
  2000.
\newblock Simultaneous Japenese-English Interpretation Based on Early
  Predictoin of English Verb.
\newblock In \emph{Proceedings of the 4th Symposium on Natural Languauge
  Processing(SNLP-2000)}, 268--273.

\bibitem[{Novitasari et~al.(2019)Novitasari, Tjandra, Sakti, and
  Nakamura}]{Novitasari2019}
Novitasari, S.; Tjandra, A.; Sakti, S.; and Nakamura, S. 2019.
\newblock {Sequence-to-Sequence Learning via Attention Transfer for Incremental
  Speech Recognition} 3835--3839.

\bibitem[{Oda et~al.(2015)Oda, Neubig, Sakti, Toda, and
  Nakamura}]{oda-etal-2015-syntax}
Oda, Y.; Neubig, G.; Sakti, S.; Toda, T.; and Nakamura, S. 2015.
\newblock Syntax-based Simultaneous Translation through Prediction of Unseen
  Syntactic Constituents.
\newblock In \emph{Proceedings of the 53rd Annual Meeting of the Association
  for Computational Linguistics and the 7th International Joint Conference on
  Natural Language Processing (Volume 1: Long Papers)}, 198--207. Beijing,
  China: Association for Computational Linguistics.
\newblock \doi{10.3115/v1/P15-1020}.
\newblock \urlprefix\url{https://www.aclweb.org/anthology/P15-1020}.

\bibitem[{Ott et~al.(2019)Ott, Edunov, Baevski, Fan, Gross, Ng, Grangier, and
  Auli}]{ott-etal-2019-fairseq}
Ott, M.; Edunov, S.; Baevski, A.; Fan, A.; Gross, S.; Ng, N.; Grangier, D.; and
  Auli, M. 2019.
\newblock fairseq: A Fast, Extensible Toolkit for Sequence Modeling.
\newblock In \emph{Proceedings of the 2019 Conference of the North {A}merican
  Chapter of the Association for Computational Linguistics (Demonstrations)},
  48--53. Minneapolis, Minnesota: Association for Computational Linguistics.
\newblock \doi{10.18653/v1/N19-4009}.
\newblock \urlprefix\url{https://www.aclweb.org/anthology/N19-4009}.

\bibitem[{Papineni et~al.(2002)Papineni, Roukos, Ward, and
  Zhu}]{papineni-etal-2002-bleu}
Papineni, K.; Roukos, S.; Ward, T.; and Zhu, W.-J. 2002.
\newblock {B}leu: a Method for Automatic Evaluation of Machine Translation.
\newblock In \emph{Proceedings of the 40th Annual Meeting of the Association
  for Computational Linguistics}, 311--318. Philadelphia, Pennsylvania, USA:
  Association for Computational Linguistics.
\newblock \doi{10.3115/1073083.1073135}.
\newblock \urlprefix\url{https://www.aclweb.org/anthology/P02-1040}.

\bibitem[{Ravanelli, Serdyuk, and Bengio(2018)}]{Ravanelli2018}
Ravanelli, M.; Serdyuk, D.; and Bengio, Y. 2018.
\newblock {Twin Regularization for online speech recognition}.
\newblock \emph{Proceedings of the Annual Conference of the International
  Speech Communication Association, INTERSPEECH} 2018-Septe(1): 3718--3722.
\newblock ISSN 19909772.
\newblock \doi{10.21437/Interspeech.2018-1407}.

\bibitem[{Sennrich, Haddow, and Birch(2016)}]{sennrich-etal-2016-neural}
Sennrich, R.; Haddow, B.; and Birch, A. 2016.
\newblock Neural Machine Translation of Rare Words with Subword Units.
\newblock In \emph{Proceedings of the 54th Annual Meeting of the Association
  for Computational Linguistics (Volume 1: Long Papers)}, 1715--1725. Berlin,
  Germany: Association for Computational Linguistics.
\newblock \doi{10.18653/v1/P16-1162}.
\newblock \urlprefix\url{https://www.aclweb.org/anthology/P16-1162}.

\bibitem[{Siahbani et~al.(2018)Siahbani, Shavarani, Alinejad, and
  Sarkar}]{siahbani-etal-2018-simultaneous}
Siahbani, M.; Shavarani, H.; Alinejad, A.; and Sarkar, A. 2018.
\newblock Simultaneous Translation using Optimized Segmentation.
\newblock In \emph{Proceedings of the 13th Conference of the Association for
  Machine Translation in the {A}mericas (Volume 1: Research Papers)}, 154--167.
  Boston, MA: Association for Machine Translation in the Americas.
\newblock \urlprefix\url{https://www.aclweb.org/anthology/W18-1815}.

\bibitem[{Vaswani et~al.(2017)Vaswani, Shazeer, Parmar, Uszkoreit, Jones,
  Gomez, Kaiser, and Polosukhin}]{NIPS2017_7181}
Vaswani, A.; Shazeer, N.; Parmar, N.; Uszkoreit, J.; Jones, L.; Gomez, A.~N.;
  Kaiser, L.~u.; and Polosukhin, I. 2017.
\newblock Attention is All you Need.
\newblock In Guyon, I.; Luxburg, U.~V.; Bengio, S.; Wallach, H.; Fergus, R.;
  Vishwanathan, S.; and Garnett, R., eds., \emph{Advances in Neural Information
  Processing Systems 30}, 5998--6008. Curran Associates, Inc.
\newblock
  \urlprefix\url{http://papers.nips.cc/paper/7181-attention-is-all-you-need.pdf}.

\bibitem[{Zhang, Xiong, and Su(2018)}]{zhang-etal-2018-accelerating}
Zhang, B.; Xiong, D.; and Su, J. 2018.
\newblock Accelerating Neural Transformer via an Average Attention Network.
\newblock In \emph{Proceedings of the 56th Annual Meeting of the Association
  for Computational Linguistics (Volume 1: Long Papers)}, 1789--1798.
  Melbourne, Australia: Association for Computational Linguistics.
\newblock \doi{10.18653/v1/P18-1166}.
\newblock \urlprefix\url{https://www.aclweb.org/anthology/P18-1166}.

\bibitem[{Zheng et~al.(2019{\natexlab{a}})Zheng, Zheng, Ma, and
  Huang}]{Zheng2019b}
Zheng, B.; Zheng, R.; Ma, M.; and Huang, L. 2019{\natexlab{a}}.
\newblock Simpler and Faster Learning of Adaptive Policies for Simultaneous
  Translation.
\newblock In \emph{Proceedings of the 2019 Conference on Empirical Methods in
  Natural Language Processing and the 9th International Joint Conference on
  Natural Language Processing (EMNLP-IJCNLP)}, 1349--1354. Hong Kong, China:
  Association for Computational Linguistics.
\newblock \doi{10.18653/v1/D19-1137}.
\newblock \urlprefix\url{https://www.aclweb.org/anthology/D19-1137}.

\bibitem[{Zheng et~al.(2019{\natexlab{b}})Zheng, Zheng, Ma, and
  Huang}]{Zheng2019a}
Zheng, B.; Zheng, R.; Ma, M.; and Huang, L. 2019{\natexlab{b}}.
\newblock Simultaneous Translation with Flexible Policy via Restricted
  Imitation Learning.
\newblock In \emph{Proceedings of the 57th Annual Meeting of the Association
  for Computational Linguistics}, 5816--5822. Florence, Italy: Association for
  Computational Linguistics.
\newblock \doi{10.18653/v1/P19-1582}.
\newblock \urlprefix\url{https://www.aclweb.org/anthology/P19-1582}.

\end{thebibliography}

\end{document}